\documentclass[sigconf]{acmart}

\usepackage{multirow}
\usepackage{caption}
\usepackage{subcaption}
\usepackage{bbm}

\newcommand{\sDelayMax}{\ensuremath{{d_{max}}}}
\newcommand{\sLogData}{\ensuremath{{\mathcal{D}}}}

\newcommand{\sProb}[1]{\ensuremath{\Pr \left( #1 \right)}}

\copyrightyear{2021}
\acmYear{2021}
\setcopyright{acmcopyright}\acmConference[SIGIR '21]{Proceedings of the 44th International ACM SIGIR Conference on Research and Development in Information Retrieval}{July 11--15, 2021}{Virtual Event, Canada}
\acmBooktitle{Proceedings of the 44th International ACM SIGIR Conference on Research and Development in Information Retrieval (SIGIR '21), July 11--15, 2021, Virtual Event, Canada}
\acmPrice{15.00}
\acmDOI{10.1145/3404835.3463045}
\acmISBN{978-1-4503-8037-9/21/07}

\begin{document}
\fancyhead{}

\title{Follow the Prophet: Accurate Online Conversion Rate Prediction in the Face of Delayed Feedback}


\author{
  Haoming Li$^{1,2}$, Feiyang Pan$^{1,2}$, Xiang Ao$^{*1,2}$, Zhao Yang$^{1,2}$, \\
  Min Lu$^3$, Junwei Pan$^3$, Dapeng Liu$^3$, Lei Xiao$^3$, Qing He$^{1,2}$
}
\affiliation{
  \institution{
    $^1$ Key Lab of Intelligent Information Processing of Chinese Academy of Sciences (CAS), \\
    Institute of Computing Technology, CAS, Beijing 100190, China \\
    $^2$ University of Chinese Academy of Sciences, Beijing 100049, China \\
    $^3$ Tencent, Shenzhen, China
  }
  \country{}
}
\email{
  {lihaoming19s,panfeiyang,aoxiang,yangzhao20s,heqing}@ict.ac.cn,
  {alfredolv,jonaspan,rocliu,shawnxiao}@tencent.com
}
\thanks{$^*$Correspondence to: Xiang Ao <aoxiang@ict.ac.cn>.}

\renewcommand{\shortauthors}{Li, et al.}

\begin{abstract}
The delayed feedback problem is one of the imperative challenges in online advertising, which is caused by the highly diversified feedback delay of a conversion varying from a few minutes to several days. 
It is hard to design an appropriate online learning system under these non-identical delay for different types of ads and users. 
In this paper, we propose to tackle the delayed feedback problem in online advertising by ``\underline{F}ollowing \underline{t}he \underline{P}rophet''~(FTP for short). The key insight is that, if the feedback came instantly for all the logged samples, we could get a model without delayed feedback, namely the ``prophet''.
Although the prophet cannot be obtained during online learning, we show that we could predict the prophet's predictions by an aggregation policy on top of a set of multi-task predictions, where each task captures the feedback patterns of different periods.
We propose the objective and optimization approach for the policy, and use the logged data to imitate the prophet. 
Extensive experiments on three real-world advertising datasets show that our method outperforms the previous state-of-the-art baselines.
\end{abstract}



\maketitle

\subsection*{}
{\fontsize{8pt}{8pt} \selectfont
  \vspace{-5mm}
  \textbf{ACM Reference Format:}\\
  Haoming Li, Feiyang Pan, Xiang Ao, Zhao Yang, Min Lu, Junwei Pan, Dapeng Liu, Lei Xiao, Qing He. 2021. Follow the Prophet: Accurate Online Conversion Rate Prediction in the Face of Delayed Feedback. In \textit{Proceedings of the 44th International ACM SIGIR Conference on Research and Development in Information Retrieval (SIGIR '21), July 11--15, 2021, Virtual Event, Canada.} ACM, New York, NY, USA, 5 pages. https://doi.org/10.1145/3404835.3463045}

\section{Introduction}

Conversion rate~(CVR) prediction is one of the most essential tasks in online advertising, which predicts the probability of whether a user will accomplish certain goals after viewing an ad, such as order placement~\cite{agarwal2010estimating,lee2012estimating,lu2017practical,pan2019predicting}. For such a task, the delayed feedback problem is one of the imperative challenges. After an ad is shown to or clicked by a user, the system can receive the conversion feedback with a delay ranging from a few minutes to several days, if there is one \cite{he2016fast}. For those ad impressions that do not have feedback yet, it is hard for the system to tell whether they will convert in the future or not. In this case, labelling these samples as negative may introduce bias to CVR prediction models. 
A customary practice is to wait for a period before assigning negative labels~\cite{he2014practical,chapelle2014modeling,ktena2019addressing}, but it makes the model hard to catch up with the most recent data.

In online advertising, since there are always a number of samples whose labels are undetermined, it is difficult to use a standard objective function (e.g., binary cross-entropy) to train the CVR prediction model. So, we consider two questions: 1) How to design an appropriate learning objective in online advertising under delayed feedback, and 2) How to learn a CVR predictor that optimizes such objective. 

In this paper, we propose a general methodology for the delayed feedback problem, named ``\underline{F}ollow \underline{t}he \underline{P}rophet'' (FTP). Our insight is, we should first construct a prophet that knows how to use the data properly so as to define the learning objective, and then learn the actual model by imitating the prophet. 

The ``prophet'' is a model trained over an idealized dataset, which is constructed to eliminate the issue of delayed feedback. We show that such a model as well as the constructed dataset can be easily obtained using a subset of historical data with complete feedback.

Second, we propose to use multiple tasks and objective functions to characterize the conversion feedback under different delayed periods. In such case, the ``prophet'' could be the optimization goal for cherry-picking the proper task for each sample. Hence, our objective is to aggregate these tasks to imitate the idealized prophet.

Finally, to accomplish that, we leverage an aggregation policy on top of the predictions of each task to yield an appropriate predictions for each sample. This aggregation policy is optimized in a well-designed training pipeline to minimize the error with an artificial prophet. This ultimately achieves our objective, ``behaving like a prophet''.

Compared to previous works, our method does not rely on specific assumptions, neither the delay distribution nor the true labels distribution. Instead, it is optimized in an objective-oriented way, by ``following the prophet.'' 
Our solution is naturally suitable for online training and easy to deploy in real-word advertising systems.

We conduct extensive experiments on three real-world advertising conversion rate prediction datasets, including two public datasets and one large-scale in-house dataset. We use a streaming simulation for evaluation to reflect the performance of each method in the online advertising system. Experimental results show that our method outperforms the other state-of-the-arts in previous works.

\section{Related Works}

Conversion prediction plays a critical role in online advertising~\cite{agarwal2010estimating,lee2012estimating,lu2017practical,pan2019predicting}, and there are several works on resolving the feedback delay issue. 
\citet{chapelle2014modeling} proposed to apply survival time analysis 
with a delayed feedback model~(DFM) to estimate the delay under an assumption of exponential distribution. \citet{yoshikawa2018nonparametric} extended DFM to a non-parametric 
model~(NoDeF). \citet{yasui2020feedback} regarded the delayed feedback as a data shift and proposed to use an importance weighting approach~(FSIW) to handle different distribution between test data and observed data.

One of the disadvantages of these methods is that they are hard to apply in gradient-descent-based online training~\cite{bartlett2007adaptive,liu2017pbodl,sahoo2018online}. 
\citet{ktena2019addressing} proposed to use a ``fake negative'' data pipeline in online streaming: all samples are assigned with a fake negative label without delay. If the system receives a conversion feedback later, they 
just \emph{duplicate} that sample with a positive label. To estimate conversion rate under this form of data, they proposed to use positive–unlabeled loss~(PU) \cite{plessis2014analysis,du2015convex,kiryo2017positive}, fake negative weighted loss~(FNW) and fake negative calibration approach~(FNC) \cite{ktena2019addressing}.

\section{Problem Formulation}

First, a CVR prediction problem without considering the delayed feedback can be formulated as probabilistic predictions of binary classification over a dataset $\sLogData=\left\{ (x_i, y_{i}) \right\}_{i=1}^N$, where each instance has input features (including ad and user attributes) $x$ and a binary label $y$ indicating whether there is a conversion~\cite{lee2012estimating,pan2020field}.

Next, we formulate an online CVR prediction problem with delayed feedback by adding a few notations. During online learning, at time $\tau$, the system maintains a dataset:
\begin{equation}
    \sLogData_\tau=\left\{ (s_i, t_i, x_i, y_{i,\tau-s_i}) \right\}_{i=1}^{N_\tau}
    \label{eq:log-data-tau}
\end{equation}
where each instance has $s$ the timestamp when it is initially logged, and $t$ the timestamp when receiving its conversion feedback. $t:=\infty$ if there is no feedback until $\tau$, and the observed label $y_{\tau-s}$ with a period of $d=\tau-s$ is defined as:
\begin{equation}
    y_d = y_{\tau-s} = \left\{
    \begin{array}{ll}
        1, & \text{if there is a conversion before } \tau, \\
        0, & \text{otherwise.}
    \end{array}
    \right.
    \label{eq:yd-ytaus}
\end{equation}

Moreover, we define $\sDelayMax$ as the maximum feedback delay, i.e., an instance initialized at time $s$ will be considered having no conversion if there is no feedback until $s + \sDelayMax$:

\begin{equation}
    y_{d'} \equiv y_{\sDelayMax}, \ \forall d' > \sDelayMax
    \label{eq:y-d-equal-y-dmax}
\end{equation}

Finally, the conversion rate can be defined as:
\begin{equation}
    \sProb{\text{conversion} | x} := \sProb{y_\sDelayMax=1 | x}
    \label{eq:pcvr-d-inf-max}
\end{equation}

\section{Follow the Prophet}

\begin{figure*}
    \includegraphics[width=\textwidth]{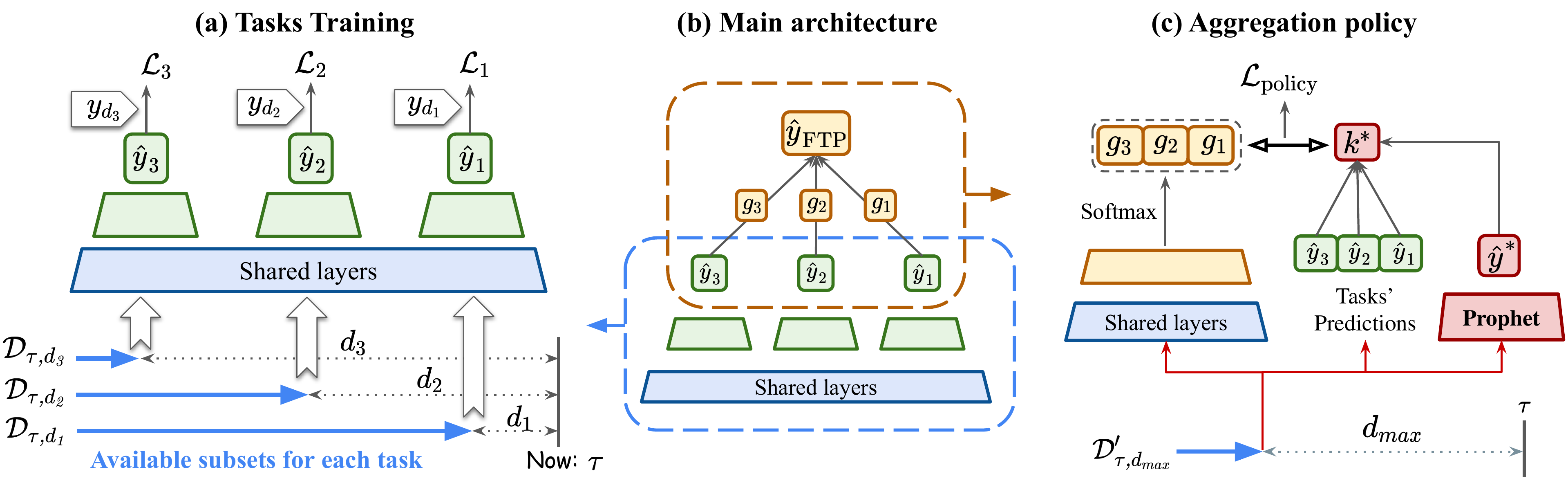}
    \caption{Design of the FTP Model and Data Pipeline.}
    \label{fig:model}
\end{figure*}

The outline of our model and data pipeline for FTP is shown in Fig.~\ref{fig:model}. 
In the following sections we will illustrate three essential components in FTP: the prophet that can give prophecies, the multiple tasks for capturing conversion feedback patterns in different periods, and the policy optimized towards the prophet to aggregate these tasks.

\subsection{Construct the prophet}
\label{sec:prophet}

As we describe in the introduction, a ``prophet'' is an ideal model that can eliminate the issue of delayed feedback. The 
issue is that, when a new instance is logged  
, we only know its features $x$, but does not have the feedback to determine the label.

So formally, at any time $\tau$, the $y_{\tau-s}$ in logging data $\sLogData_\tau$ is not the label of CVR prediction problem (which need to be~$y_{\sDelayMax}$ in Eq.~\ref{eq:pcvr-d-inf-max}).

An ideal CVR prediction model could be trained if we could get a set of ideal data with all the feedback, which would be:

\begin{equation}
    \sLogData^*_\tau = \left\{ (s_i, t_i, x_i, y_{\sDelayMax}) \right\}_{i=1}^{N_\tau}
\end{equation}

We define that, at any time $\tau$:

\medskip
A \emph{prophet} is a well-trained CVR model based on $D^*_\tau$.

A \emph{prophecy} is the prediction of the \emph{prophet}.
\medskip

Obviously, we cannot obtain $\sLogData^*_\tau$ if the time now is $\tau$. However, according to Eq.~\ref{eq:y-d-equal-y-dmax}, on a subset of $\sLogData_\tau$ with the instances subject to $s<\tau - \sDelayMax$, we have:

\begin{equation}
    \sLogData_{\tau, \sDelayMax} := \sLogData_{\tau}|_{s_i<\tau-\sDelayMax} \equiv \sLogData^*_{\tau - \sDelayMax}
    \label{eq:log-dmax-equal-log-star}
\end{equation}

$\sLogData_{\tau,\sDelayMax}$ contains instances with complete feedback until $\tau$, since by the definition of $\sDelayMax$, all the feedback in this subset should already be received. Therefore, 
it is possible to construct a prophet until $\tau - \sDelayMax$ with logging data subset $\sLogData_{\tau,\sDelayMax}$.

\subsection{Multi-task learning for delayed feedback}
\label{sec:multiple-tasks}

Recall that, without considering the delayed feedback, we can use the standard binary cross-entropy objective function to optimize a single CVR prediction model. However, when under delayed feedback, this kind of optimization approach fails to properly handle samples without feedback yet.

Therefore, instead of optimize the CVR prediction directly, we propose to set up multiple tasks and objective functions to characterize different delayed periods:

\begin{center}
    Task $k$: \quad $\hat{y}_k = f_k(x) = \sProb { y_{d_k}=1 | x }$, \quad $k = 1, 2, ..., K$
\end{center}
We assume that $0 < d_1 < d_2 < ... < d_K = \sDelayMax$. 

Similar to 
Eq.~\ref{eq:log-dmax-equal-log-star}, for each task $k$, the available logging subset is:
\begin{equation}
    \sLogData_{\tau, d_k} := \sLogData_\tau|_{s_i<\tau - d_k}
\end{equation}

The loss function for each task $k$ is the binary cross-entropy
\begin{equation}
    \mathcal{L}_k = - y_{d_k} \log(f_k(x)) - (1 - y_{d_k}) \log(1 - f_k(x))
\end{equation}

By limited the training data to a subset subject to $s<\tau - d_k$, the label $y_{d_k}$ is always available at time $\tau$. So, all the samples are entirely labeled during training. This is one of the major differences between our method and previous works. 

Another advantage is that it is easy to build data pipelines for online training, as shown in Fig.~\ref{fig:model}(a). Each task can fetch new available training data continuously from its own data pipeline, and perform incremental updates.

Various forms of DNN-based multi-task networks can be applied to the model~\cite{baxter2000model,ruder2017overview}, including the widely used shared-bottom structure~\cite{caruana1997multitask}, cross-lingual parameter sharing~\cite{duong2015low}, MoE~\cite{jacobs1991adaptive,eigen2013learning,shazeer2017outrageously},  MMoE~\cite{ma2018modeling} and MT-FwFM~\cite{pan2019predicting}, etc.

\subsection{Aggregation policy to imitate the prophet}
Now, we introduce the aggregation policy to imitate the prophet, which combines these tasks to make the final prediction. 

Our aggregation policy is a model that outputs a set of weights (by a softmax activation) for the tasks, denoted by $(g_1(x), \dots, g_K(x))$ for a given sample with input $x$. While making prediction (c.f. Fig.~\ref{fig:model}(b)), we use the weighted average of the tasks, i.e.,
\begin{equation}
    \hat y_{\mathrm{FTP}} = \sum_{k=1}^K g_k(x) f_k(x)
\end{equation}

\subsubsection{Policy loss}

The optimization approach for the policy model is exactly as the name of our method: follow the prophet, as shown in Fig.~\ref{fig:model}(c).

Since we have all the models, we can have the predictions of the tasks $\hat y_k=f_k(x)$, $k=1,\dots,K$, as well as the prediction of the prophet $y^*=f^*(x)$. Then, we can find the task $k^*$ which has the closest prediction to the prophecy,
\begin{equation}
    k^* = \mathop{\arg\min}_k \left| \hat{y}^* - \hat{y}_{k} \right|.
    \label{eq:k-star}
\end{equation}

We want the policy to assign the largest weight to the task $k^*$, so the loss on the sample is an ordinary cross-entropy
\begin{equation}
    \mathcal{L}_\text{policy} = -\sum_{k=1}^K \mathbbm{1}(k=k^*) \log\left(g_k(x)\right)
\end{equation}

\subsubsection{Training data pipeline}

In order to compute Eq.~\ref{eq:k-star}, we save the tasks' prediction into the online logs, and we denote this extended online logs as:

\begin{equation}
    \sLogData'_\tau = \left\{ (s_i, x_i, y_{\tau-s_i}, \hat{y}_{\text{tasks},i}) \right\}_{i=1}^{N_\tau}
\end{equation}
where $\hat{y}_{\text{tasks},i} = (\hat{y}_{1,i}, \hat{y}_{2,i}, ..., \hat{y}_{K,i})$ is the \emph{online} predictions of multiple tasks.

As we discuss in the previous sections, the prophet can only be constructed on $\sLogData_{\tau,\sDelayMax}$, so the training of aggregation policy should also based on the subset of the extended logs $\sLogData'_{\tau,\sDelayMax}$. 

\section{Experiments}

\begin{figure*}[t]
    \captionsetup{justification=centering}

    \begin{minipage}[b]{0.78\textwidth}
        \includegraphics[width=\linewidth]{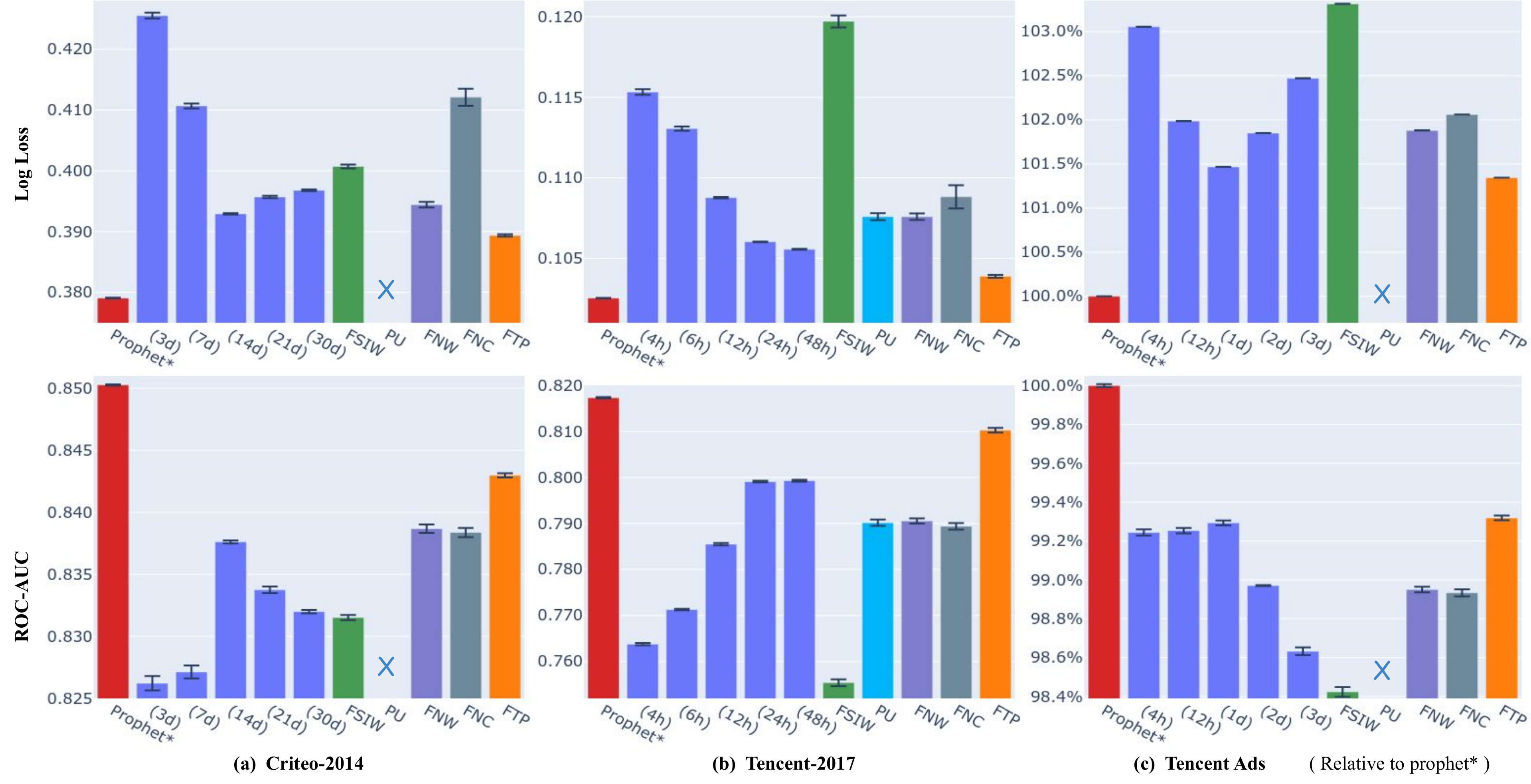}
        \vspace{-6mm}
        \caption{Results. The methods ``($d$)'' means ``Waiting($d$)'' for short.}
        \label{fig:results}
    \end{minipage}
    \hfill
    \begin{minipage}[b]{0.21\textwidth}
        \centering
        \begin{subtable}{\linewidth}
        \centering
        \begin{tabular}{lrr}
        \toprule
        \small{$d_k$}  & \small{ Feedback\%} & \small{Best}\% \\
        \midrule
           1d &           60\% &             9\% \\
           7d &           80\% &            18\% \\
          14d &           90\% &            24\% \\
          21d &           95\% &            24\% \\
          30d &          100\% &            25\% \\
        \bottomrule
        \end{tabular}
        \caption{Criteo-2014}
        \end{subtable}
        \\
        \begin{subtable}{\linewidth}
        \centering
        \begin{tabular}{lrr}
        \toprule
        \small{$d_k$}  & \small{ Feedback\%} & \small{Best}\% \\
        \midrule
           1h &           73\% &            23\% \\
           6h &           83\% &            23\% \\
          24h &           95\% &            26\% \\
          48h &          100\% &            28\% \\
        \bottomrule
        \end{tabular}
        \caption{Tencent-2017}
        \end{subtable}
        \vspace{2mm}
        \captionof{table}{Feedback rate for $d<d_k$ and frequency of task $k$ being the best task.}
        \label{tab:best-tasks}
    \end{minipage}
\end{figure*}

\subsection{Streaming simulation set-up}

We focus on the online system and use a simulation of online streaming while performing offline evaluation. The original dataset is $\sLogData^*$. A simulated ``current'' timestamp $\tau$ is used to control the evaluation procedure. During the training process, only $\sLogData_\tau$ is available. Every time after the model update, it will be used to predict and evaluate on the instances in next hour. Finally, the timestamp $\tau$ moves to one hour later, and we repeat the steps above.

\subsection{Datasets}

\subsubsection*{Criteo Conversion Logs}\footnote{https://labs.criteo.com/2013/12/conversion-logs-dataset/} (Criteo-2014). This is a widely used public dataset for the delayed feedback problem~\cite{chapelle2014modeling}. It includes 60 days data, with $\sDelayMax$ as 30 days. The last 20 days are used for evaluation.

\subsubsection*{Tencent Advertising Algorithm Competition 2017}\footnote{https://algo.qq.com/?lang=en} (Tencent-2017). This 
dataset includes 14 days of click and conversion logs. According to the discussion about this competition, we set $\sDelayMax=$ to 2 days, and data of last 2 days are excluded. The last remaining 5 days are used for evaluation.

\subsubsection*{Tencent Ads Conversion Logs} (Tencent Ads). To verify the effectiveness of our method on production data, we collect 3 months of conversion logs from Tencent Ads, with about 200 million instances after sampling. We use the instances in the last month for evaluation.

\subsection{Compared methods}

\noindent \textbf{Prophet}$^*$: An ideal CVR model constructed on $\sLogData^*$ for evaluation.

\noindent \textbf{Waiting}($d$): A method that only updates on $\sLogData_{\tau,d}$ $(s<\tau-d)$.

\noindent \textbf{FSIW}: Feedback shift importance weighting \cite{yasui2020feedback}. Note that this method is not originally designed for streaming settings.

\noindent \textbf{PU}, \textbf{FNW}, \textbf{FNC}: Using a fake negative data pipeline with a PU loss / FNW loss / FN Calibration.\cite{ktena2019addressing}.

\noindent \textbf{FTP}: Our method. We set the tasks to (1~day, 7~days, 14~days, 21~days, 30~days) on Criteo-2014, (1~hour, 6~hours, 24~hours, 48~hours) on Tencent-2017, and (4~hours, 1~day, 5~days) on Tencent Ads.

\medskip

We use a DNN model as the base model for all the methods. For the two public datasets, all features are transformed into embeddings, and concatenated together to feed into two dense layers with 128 units in each. The activation function is Leaky ReLU. L2 regularization is set to $10^{-6}$ on Criteo-2014 and $10^{-7}$ on Tencent-2017.
The models are update by the Adam optimizer~\cite{kingma2014adam}. In our FTP method, all the tasks and the policy network share the embedding layers and the first dense layer at the bottom\footnote{Our code on the public datasets: https://github.com/kmdict/FollowTheProphet}.

\subsection{Results}

We conduct 5 random runs on the two public datasets, and the results are shown in Fig.~\ref{fig:results}~(a,b). Results on in-house Tencent Ads data is shown in Fig.~\ref{fig:results}~(c), and the relative metrics compared to prophet$^*$ are reported. 

From Fig.~\ref{fig:results}, we observe that our FTP outperforms all other methods stably on all datasets.
The performance of FSIW is not satisfactory, because it is not design for streaming settings, and fail to make use of recent feedback directly. 
PU loss is hard to converge on Criteo-2014 and Tencent Ads in our experiments. A similar issue is also mentioned in \cite{ktena2019addressing}. 

Over all the datasets, we find that the \textit{Waiting} method with a well-tuned waiting period $d$ can also have considerable performance. There isn't much mentioned about this baseline in previous works. Our FTP is the only method that outperforms the best-tuned \textit{Waiting} baseline.

To further analyze the success of FTP, we show statistics about different tasks in FTP in Table.~\ref{tab:best-tasks}. ``Feedback\%'' indicates how many percentages of feedback can be received within the specific task's period. ``Best\%'' counts up how many times that each task becomes the ``best'' task, with a prediction closest to the prophecy. 

We highlight two characteristics that may lead to the success of FTP. First, all the tasks have considerable opportunity to become the best task, which indicates that they can capture different patterns under different delay periods. Second, although there is already 60\% of feedback can be received within 1 day on Criteo-2014, it obviously does \textit{not} mean the task with a delay of 1 day should always be the best task among them. FTP does not learn from the unknown delay time distribution directly, but use a more objective-oriented way: following the prophet.

\section{Conclusion}

In this paper, we focus on the online CVR prediction and propose ``Following the Prophet''~(FTP) method to tackle the delayed feedback problem. We construct an aggregation policy on top of multi-task predictions, where each task captures the feedback pattern during different periods. The policy is optimized towards an ideal ``prophet'' based on historical logged data. Extensive experiments on three real-world advertising datasets show that our method outperforms previous state-of-the-arts.

\begin{acks}
This work is sponsored by CCF-Tencent Open Fund. The research work is also supported by the National Key Research and Development Program of China under Grant No. 2017YFB1002104, the National Natural Science Foundation of China under Grant No. 92046003, 61976204, U1811461. Xiang Ao is also supported by the Project of Youth Innovation Promotion Association CAS and Beijing Nova Program Z201100006820062.
\end{acks}

\bibliographystyle{ACM-Reference-Format}
\bibliography{base}


\begin{thebibliography}{26}


\ifx \showCODEN    \undefined \def \showCODEN     #1{\unskip}     \fi
\ifx \showDOI      \undefined \def \showDOI       #1{#1}\fi
\ifx \showISBNx    \undefined \def \showISBNx     #1{\unskip}     \fi
\ifx \showISBNxiii \undefined \def \showISBNxiii  #1{\unskip}     \fi
\ifx \showISSN     \undefined \def \showISSN      #1{\unskip}     \fi
\ifx \showLCCN     \undefined \def \showLCCN      #1{\unskip}     \fi
\ifx \shownote     \undefined \def \shownote      #1{#1}          \fi
\ifx \showarticletitle \undefined \def \showarticletitle #1{#1}   \fi
\ifx \showURL      \undefined \def \showURL       {\relax}        \fi
\providecommand\bibfield[2]{#2}
\providecommand\bibinfo[2]{#2}
\providecommand\natexlab[1]{#1}
\providecommand\showeprint[2][]{arXiv:#2}

\bibitem[\protect\citeauthoryear{Agarwal, Agrawal, Khanna, and Kota}{Agarwal
  et~al\mbox{.}}{2010}]%
        {agarwal2010estimating}
\bibfield{author}{\bibinfo{person}{Deepak Agarwal}, \bibinfo{person}{Rahul
  Agrawal}, \bibinfo{person}{Rajiv Khanna}, {and} \bibinfo{person}{Nagaraj
  Kota}.} \bibinfo{year}{2010}\natexlab{}.
\newblock \showarticletitle{Estimating rates of rare events with multiple
  hierarchies through scalable log-linear models}. In
  \bibinfo{booktitle}{\emph{KDD}}. \bibinfo{pages}{213--222}.
\newblock


\bibitem[\protect\citeauthoryear{Bartlett, Hazan, and Rakhlin}{Bartlett
  et~al\mbox{.}}{2007}]%
        {bartlett2007adaptive}
\bibfield{author}{\bibinfo{person}{Peter~L Bartlett}, \bibinfo{person}{Elad
  Hazan}, {and} \bibinfo{person}{Alexander Rakhlin}.}
  \bibinfo{year}{2007}\natexlab{}.
\newblock \showarticletitle{Adaptive online gradient descent}. In
  \bibinfo{booktitle}{\emph{NeurIPS}}. \bibinfo{pages}{65--72}.
\newblock


\bibitem[\protect\citeauthoryear{Baxter}{Baxter}{2000}]%
        {baxter2000model}
\bibfield{author}{\bibinfo{person}{Jonathan Baxter}.}
  \bibinfo{year}{2000}\natexlab{}.
\newblock \showarticletitle{A model of inductive bias learning}.
\newblock \bibinfo{journal}{\emph{JAIR}}  \bibinfo{volume}{12}
  (\bibinfo{year}{2000}), \bibinfo{pages}{149--198}.
\newblock


\bibitem[\protect\citeauthoryear{Caruana}{Caruana}{1997}]%
        {caruana1997multitask}
\bibfield{author}{\bibinfo{person}{Rich Caruana}.}
  \bibinfo{year}{1997}\natexlab{}.
\newblock \showarticletitle{Multitask learning}.
\newblock \bibinfo{journal}{\emph{Machine learning}} \bibinfo{volume}{28},
  \bibinfo{number}{1} (\bibinfo{year}{1997}), \bibinfo{pages}{41--75}.
\newblock


\bibitem[\protect\citeauthoryear{Chapelle}{Chapelle}{2014}]%
        {chapelle2014modeling}
\bibfield{author}{\bibinfo{person}{Olivier Chapelle}.}
  \bibinfo{year}{2014}\natexlab{}.
\newblock \showarticletitle{Modeling delayed feedback in display advertising}.
  In \bibinfo{booktitle}{\emph{KDD}}. \bibinfo{pages}{1097--1105}.
\newblock


\bibitem[\protect\citeauthoryear{Du~Plessis, Niu, and Sugiyama}{Du~Plessis
  et~al\mbox{.}}{2015}]%
        {du2015convex}
\bibfield{author}{\bibinfo{person}{Marthinus Du~Plessis}, \bibinfo{person}{Gang
  Niu}, {and} \bibinfo{person}{Masashi Sugiyama}.}
  \bibinfo{year}{2015}\natexlab{}.
\newblock \showarticletitle{Convex formulation for learning from positive and
  unlabeled data}. In \bibinfo{booktitle}{\emph{ICML}}.
  \bibinfo{pages}{1386--1394}.
\newblock


\bibitem[\protect\citeauthoryear{Duong, Cohn, Bird, and Cook}{Duong
  et~al\mbox{.}}{2015}]%
        {duong2015low}
\bibfield{author}{\bibinfo{person}{Long Duong}, \bibinfo{person}{Trevor Cohn},
  \bibinfo{person}{Steven Bird}, {and} \bibinfo{person}{Paul Cook}.}
  \bibinfo{year}{2015}\natexlab{}.
\newblock \showarticletitle{Low resource dependency parsing: Cross-lingual
  parameter sharing in a neural network parser}. In
  \bibinfo{booktitle}{\emph{ACL}}. \bibinfo{pages}{845--850}.
\newblock


\bibitem[\protect\citeauthoryear{Eigen, Ranzato, and Sutskever}{Eigen
  et~al\mbox{.}}{2013}]%
        {eigen2013learning}
\bibfield{author}{\bibinfo{person}{David Eigen}, \bibinfo{person}{Marc'Aurelio
  Ranzato}, {and} \bibinfo{person}{Ilya Sutskever}.}
  \bibinfo{year}{2013}\natexlab{}.
\newblock \showarticletitle{Learning factored representations in a deep mixture
  of experts}. In \bibinfo{booktitle}{\emph{ICLR}}.
\newblock


\bibitem[\protect\citeauthoryear{He, Pan, Jin, Xu, Liu, Xu, Shi, Atallah,
  Herbrich, Bowers, et~al\mbox{.}}{He et~al\mbox{.}}{2014}]%
        {he2014practical}
\bibfield{author}{\bibinfo{person}{Xinran He}, \bibinfo{person}{Junfeng Pan},
  \bibinfo{person}{Ou Jin}, \bibinfo{person}{Tianbing Xu}, \bibinfo{person}{Bo
  Liu}, \bibinfo{person}{Tao Xu}, \bibinfo{person}{Yanxin Shi},
  \bibinfo{person}{Antoine Atallah}, \bibinfo{person}{Ralf Herbrich},
  \bibinfo{person}{Stuart Bowers}, {et~al\mbox{.}}}
  \bibinfo{year}{2014}\natexlab{}.
\newblock \showarticletitle{Practical lessons from predicting clicks on ads at
  facebook}.
\newblock In \bibinfo{booktitle}{\emph{AdKDD}}. \bibinfo{pages}{1--9}.
\newblock


\bibitem[\protect\citeauthoryear{He, Zhang, Kan, and Chua}{He
  et~al\mbox{.}}{2016}]%
        {he2016fast}
\bibfield{author}{\bibinfo{person}{Xiangnan He}, \bibinfo{person}{Hanwang
  Zhang}, \bibinfo{person}{Min-Yen Kan}, {and} \bibinfo{person}{Tat-Seng
  Chua}.} \bibinfo{year}{2016}\natexlab{}.
\newblock \showarticletitle{Fast matrix factorization for online recommendation
  with implicit feedback}. In \bibinfo{booktitle}{\emph{SIGIR}}.
  \bibinfo{pages}{549--558}.
\newblock


\bibitem[\protect\citeauthoryear{Jacobs, Jordan, Nowlan, and Hinton}{Jacobs
  et~al\mbox{.}}{1991}]%
        {jacobs1991adaptive}
\bibfield{author}{\bibinfo{person}{Robert~A Jacobs}, \bibinfo{person}{Michael~I
  Jordan}, \bibinfo{person}{Steven~J Nowlan}, {and} \bibinfo{person}{Geoffrey~E
  Hinton}.} \bibinfo{year}{1991}\natexlab{}.
\newblock \showarticletitle{Adaptive mixtures of local experts}.
\newblock \bibinfo{journal}{\emph{Neural computation}} \bibinfo{volume}{3},
  \bibinfo{number}{1} (\bibinfo{year}{1991}), \bibinfo{pages}{79--87}.
\newblock


\bibitem[\protect\citeauthoryear{Kingma and Ba}{Kingma and Ba}{2015}]%
        {kingma2014adam}
\bibfield{author}{\bibinfo{person}{Diederik~P Kingma} {and}
  \bibinfo{person}{Jimmy Ba}.} \bibinfo{year}{2015}\natexlab{}.
\newblock \showarticletitle{Adam: A method for stochastic optimization}. In
  \bibinfo{booktitle}{\emph{ICLR}}.
\newblock


\bibitem[\protect\citeauthoryear{Kiryo, Niu, du~Plessis, and Sugiyama}{Kiryo
  et~al\mbox{.}}{2017}]%
        {kiryo2017positive}
\bibfield{author}{\bibinfo{person}{Ryuichi Kiryo}, \bibinfo{person}{Gang Niu},
  \bibinfo{person}{Marthinus~C du Plessis}, {and} \bibinfo{person}{Masashi
  Sugiyama}.} \bibinfo{year}{2017}\natexlab{}.
\newblock \showarticletitle{Positive-unlabeled learning with non-negative risk
  estimator}. In \bibinfo{booktitle}{\emph{NeurIPS}}.
  \bibinfo{pages}{1674--1684}.
\newblock


\bibitem[\protect\citeauthoryear{Ktena, Tejani, Theis, Myana, Dilipkumar,
  Husz{\'a}r, Yoo, and Shi}{Ktena et~al\mbox{.}}{2019}]%
        {ktena2019addressing}
\bibfield{author}{\bibinfo{person}{Sofia~Ira Ktena}, \bibinfo{person}{Alykhan
  Tejani}, \bibinfo{person}{Lucas Theis}, \bibinfo{person}{Pranay~Kumar Myana},
  \bibinfo{person}{Deepak Dilipkumar}, \bibinfo{person}{Ferenc Husz{\'a}r},
  \bibinfo{person}{Steven Yoo}, {and} \bibinfo{person}{Wenzhe Shi}.}
  \bibinfo{year}{2019}\natexlab{}.
\newblock \showarticletitle{Addressing delayed feedback for continuous training
  with neural networks in CTR prediction}. In
  \bibinfo{booktitle}{\emph{RecSys}}. \bibinfo{pages}{187--195}.
\newblock


\bibitem[\protect\citeauthoryear{Lee, Orten, Dasdan, and Li}{Lee
  et~al\mbox{.}}{2012}]%
        {lee2012estimating}
\bibfield{author}{\bibinfo{person}{Kuang-chih Lee}, \bibinfo{person}{Burkay
  Orten}, \bibinfo{person}{Ali Dasdan}, {and} \bibinfo{person}{Wentong Li}.}
  \bibinfo{year}{2012}\natexlab{}.
\newblock \showarticletitle{Estimating conversion rate in display advertising
  from past erformance data}. In \bibinfo{booktitle}{\emph{KDD}}.
  \bibinfo{pages}{768--776}.
\newblock


\bibitem[\protect\citeauthoryear{Liu, Xue, Xiao, and Zhang}{Liu
  et~al\mbox{.}}{2017}]%
        {liu2017pbodl}
\bibfield{author}{\bibinfo{person}{Xun Liu}, \bibinfo{person}{Wei Xue},
  \bibinfo{person}{Lei Xiao}, {and} \bibinfo{person}{Bo Zhang}.}
  \bibinfo{year}{2017}\natexlab{}.
\newblock \showarticletitle{Pbodl: Parallel bayesian online deep learning for
  click-through rate prediction in tencent advertising system}.
\newblock \bibinfo{journal}{\emph{arXiv preprint arXiv:1707.00802}}
  (\bibinfo{year}{2017}).
\newblock


\bibitem[\protect\citeauthoryear{Lu, Pan, Wang, Pan, Wan, and Yang}{Lu
  et~al\mbox{.}}{2017}]%
        {lu2017practical}
\bibfield{author}{\bibinfo{person}{Quan Lu}, \bibinfo{person}{Shengjun Pan},
  \bibinfo{person}{Liang Wang}, \bibinfo{person}{Junwei Pan},
  \bibinfo{person}{Fengdan Wan}, {and} \bibinfo{person}{Hongxia Yang}.}
  \bibinfo{year}{2017}\natexlab{}.
\newblock \showarticletitle{A practical framework of conversion rate prediction
  for online display advertising}.
\newblock In \bibinfo{booktitle}{\emph{AdKDD}}. \bibinfo{pages}{1--9}.
\newblock


\bibitem[\protect\citeauthoryear{Ma, Zhao, Yi, Chen, Hong, and Chi}{Ma
  et~al\mbox{.}}{2018}]%
        {ma2018modeling}
\bibfield{author}{\bibinfo{person}{Jiaqi Ma}, \bibinfo{person}{Zhe Zhao},
  \bibinfo{person}{Xinyang Yi}, \bibinfo{person}{Jilin Chen},
  \bibinfo{person}{Lichan Hong}, {and} \bibinfo{person}{Ed~H Chi}.}
  \bibinfo{year}{2018}\natexlab{}.
\newblock \showarticletitle{Modeling task relationships in multi-task learning
  with multi-gate mixture-of-experts}. In \bibinfo{booktitle}{\emph{KDD}}.
  \bibinfo{pages}{1930--1939}.
\newblock


\bibitem[\protect\citeauthoryear{Pan, Ao, Tang, Lu, Liu, Xiao, and He}{Pan
  et~al\mbox{.}}{2020}]%
        {pan2020field}
\bibfield{author}{\bibinfo{person}{Feiyang Pan}, \bibinfo{person}{Xiang Ao},
  \bibinfo{person}{Pingzhong Tang}, \bibinfo{person}{Min Lu},
  \bibinfo{person}{Dapeng Liu}, \bibinfo{person}{Lei Xiao}, {and}
  \bibinfo{person}{Qing He}.} \bibinfo{year}{2020}\natexlab{}.
\newblock \showarticletitle{Field-aware Calibration: A Simple and Empirically
  Strong Method for Reliable Probabilistic Predictions}. In
  \bibinfo{booktitle}{\emph{WWW}}. \bibinfo{pages}{729--739}.
\newblock


\bibitem[\protect\citeauthoryear{Pan, Mao, Ruiz, Sun, and Flores}{Pan
  et~al\mbox{.}}{2019}]%
        {pan2019predicting}
\bibfield{author}{\bibinfo{person}{Junwei Pan}, \bibinfo{person}{Yizhi Mao},
  \bibinfo{person}{Alfonso~Lobos Ruiz}, \bibinfo{person}{Yu Sun}, {and}
  \bibinfo{person}{Aaron Flores}.} \bibinfo{year}{2019}\natexlab{}.
\newblock \showarticletitle{Predicting different types of conversions with
  multi-task learning in online advertising}. In
  \bibinfo{booktitle}{\emph{KDD}}. \bibinfo{pages}{2689--2697}.
\newblock


\bibitem[\protect\citeauthoryear{Plessis, Niu, and Sugiyama}{Plessis
  et~al\mbox{.}}{2014}]%
        {plessis2014analysis}
\bibfield{author}{\bibinfo{person}{Marthinus C~du Plessis},
  \bibinfo{person}{Gang Niu}, {and} \bibinfo{person}{Masashi Sugiyama}.}
  \bibinfo{year}{2014}\natexlab{}.
\newblock \showarticletitle{Analysis of learning from positive and unlabeled
  data}. In \bibinfo{booktitle}{\emph{NeurIPS}}. \bibinfo{pages}{703--711}.
\newblock


\bibitem[\protect\citeauthoryear{Ruder}{Ruder}{2017}]%
        {ruder2017overview}
\bibfield{author}{\bibinfo{person}{Sebastian Ruder}.}
  \bibinfo{year}{2017}\natexlab{}.
\newblock \showarticletitle{An overview of multi-task learning in deep neural
  networks}.
\newblock \bibinfo{journal}{\emph{arXiv preprint arXiv:1706.05098}}
  (\bibinfo{year}{2017}).
\newblock


\bibitem[\protect\citeauthoryear{Sahoo, Pham, Lu, and Hoi}{Sahoo
  et~al\mbox{.}}{2018}]%
        {sahoo2018online}
\bibfield{author}{\bibinfo{person}{Doyen Sahoo}, \bibinfo{person}{Quang Pham},
  \bibinfo{person}{Jing Lu}, {and} \bibinfo{person}{Steven~CH Hoi}.}
  \bibinfo{year}{2018}\natexlab{}.
\newblock \showarticletitle{Online deep learning: learning deep neural networks
  on the fly}. In \bibinfo{booktitle}{\emph{IJCAI}}.
  \bibinfo{pages}{2660--2666}.
\newblock


\bibitem[\protect\citeauthoryear{Shazeer, Mirhoseini, Maziarz, Davis, Le,
  Hinton, and Dean}{Shazeer et~al\mbox{.}}{2017}]%
        {shazeer2017outrageously}
\bibfield{author}{\bibinfo{person}{Noam Shazeer}, \bibinfo{person}{Azalia
  Mirhoseini}, \bibinfo{person}{Krzysztof Maziarz}, \bibinfo{person}{Andy
  Davis}, \bibinfo{person}{Quoc Le}, \bibinfo{person}{Geoffrey Hinton}, {and}
  \bibinfo{person}{Jeff Dean}.} \bibinfo{year}{2017}\natexlab{}.
\newblock \showarticletitle{Outrageously large neural networks: The
  sparsely-gated mixture-of-experts layer}. In
  \bibinfo{booktitle}{\emph{ICLR}}.
\newblock


\bibitem[\protect\citeauthoryear{Yasui, Morishita, Komei, and Shibata}{Yasui
  et~al\mbox{.}}{2020}]%
        {yasui2020feedback}
\bibfield{author}{\bibinfo{person}{Shota Yasui}, \bibinfo{person}{Gota
  Morishita}, \bibinfo{person}{Fujita Komei}, {and} \bibinfo{person}{Masashi
  Shibata}.} \bibinfo{year}{2020}\natexlab{}.
\newblock \showarticletitle{A Feedback Shift Correction in Predicting
  Conversion Rates under Delayed Feedback}. In \bibinfo{booktitle}{\emph{WWW}}.
  \bibinfo{pages}{2740--2746}.
\newblock


\bibitem[\protect\citeauthoryear{Yoshikawa and Imai}{Yoshikawa and
  Imai}{2018}]%
        {yoshikawa2018nonparametric}
\bibfield{author}{\bibinfo{person}{Yuya Yoshikawa} {and}
  \bibinfo{person}{Yusaku Imai}.} \bibinfo{year}{2018}\natexlab{}.
\newblock \showarticletitle{A nonparametric delayed feedback model for
  conversion rate prediction}.
\newblock \bibinfo{journal}{\emph{arXiv preprint arXiv:1802.00255}}
  (\bibinfo{year}{2018}).
\newblock


\end{thebibliography}

\end{document}